\documentclass[10pt,twocolumn,letterpaper]{article}

\usepackage[pagenumbers]{cvpr} 

\definecolor{cvprblue}{rgb}{0.21,0.49,0.74}
\usepackage[pagebackref,breaklinks,colorlinks,allcolors=cvprblue]{hyperref}

\def\paperID{30} 
\def\confName{CVPR}
\def\confYear{2026}
 
\title{Efficient Sim-to-Real Transfer of World-Action Models from Synthetic Priors}
 
\author{
Zixing Wang\textsuperscript{1,2} \quad 
Kausik Sivakumar\textsuperscript{2} \quad 
Jinghuan Shang\textsuperscript{2} \quad 
Yafei Hu\textsuperscript{2} \quad 
Zhaoming Xie\textsuperscript{2} \quad \\
Ran Gong\textsuperscript{2,$\dagger$} \quad 
Xiaohan Zhang\textsuperscript{2,$\dagger$} \quad 
Karl Schmeckpeper\textsuperscript{2,$\dagger$}\\
\textsuperscript{1}Purdue University \quad 
\textsuperscript{2}Robotics and AI Institute \quad
\textsuperscript{$\dagger$} Equal Advising\\
}
 
\newcommand{\teaserwidth}{1.0\textwidth}
 
\begin{document}
 
\twocolumn[{%
  \maketitle
  \vspace{-8pt}
  \begin{center}
    \includegraphics[width=\teaserwidth]{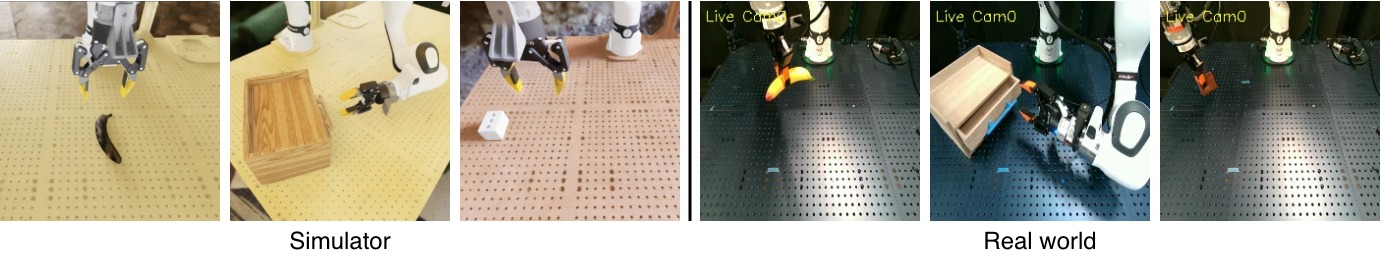}
    \vspace{2pt}\\
    {\small \textbf{Figure 1.} Real-world zero-shot RGB sim-to-real rollouts on a Franka Research 3 for three representative tasks, shown left to right: \textit{lift banana}, \textit{lift brick}, and \textit{open drawer}. The same policy family is trained only from synthetic demonstrations and deployed directly on the real robot.}
  \end{center}
  \label{fig:title}
  \vspace{1pt}
}]
\setcounter{figure}{1}
 
\begin{figure*}[t]
\centering
\includegraphics[width=0.95\linewidth]{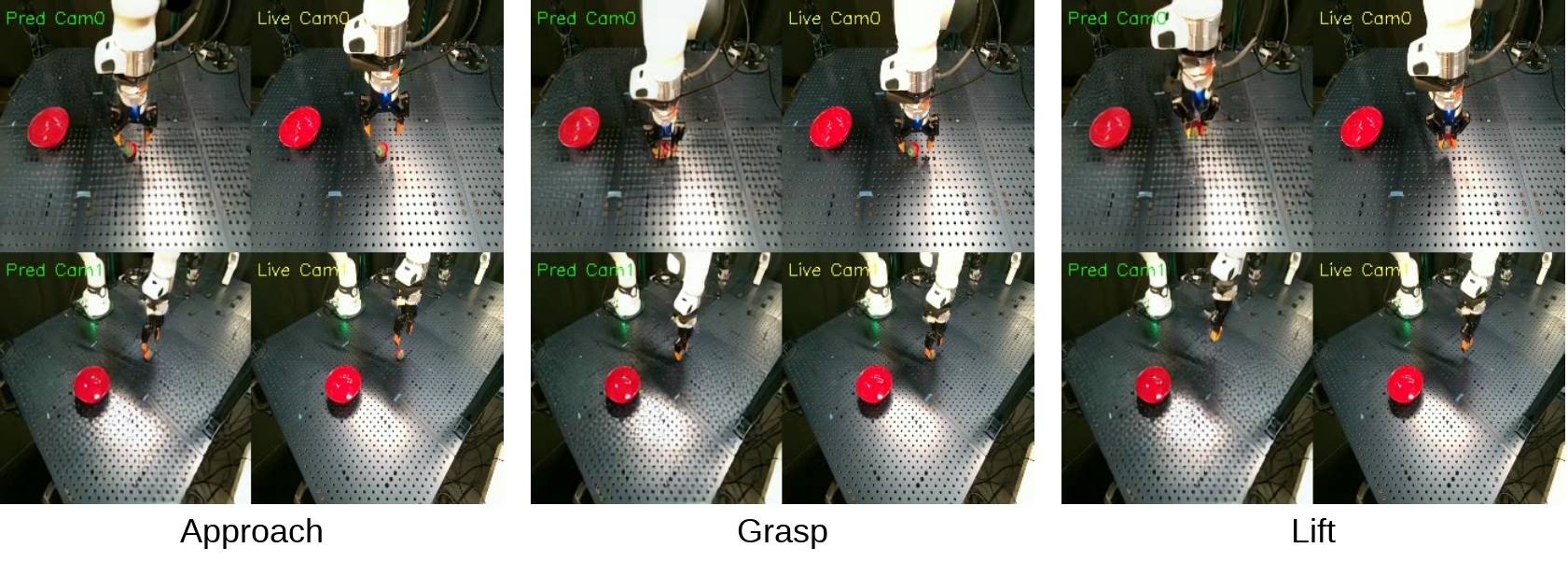}
\vspace{-6pt}
\caption{Qualitative comparison between model-predicted and synchronized live camera observations during \textit{put strawberry into bowl}. The three groups correspond to the approach, grasp, and lift stages. In each group, the overlays ``Pred Cam'' and ``Live Cam'' denote the predicted future RGB view from the world-action model and the real camera frame observed during execution, respectively. The side-by-side alignment illustrates that the policy's predicted frames preserve the object location, robot pose, and task stage despite training only in simulation.}
\label{fig:pred_effect}
\vspace{-8pt}
\end{figure*}

\begin{abstract}
Bridging the sim-to-real gap is a core challenge in deploying learned manipulation policies. Sim-to-real learning is attractive because it can replace expensive real robot demonstrations with scalable synthetic data, yet world-action models have not previously been shown to transfer from simulation to real robotic manipulation. We study whether a world-action model can be trained from synthetic priors and deployed zero-shot in the real world. To this end, we build upon Cosmos Policy, a video diffusion model adapted for visuomotor control. We construct simulation environments with extensive domain randomization and generate demonstrations using the AnyTask motion planning pipeline. We evaluate our approach across object lifting, drawer opening, and pick-and-place tasks using ${\sim}800$ synthetic demonstrations per task and no real demonstrations. When deployed zero-shot on a Franka Robot, our policy attains a 35\% average success rate. To our knowledge, this represents the first successful sim-to-real transfer of a world-action model for robotic manipulation.

~\textcolor{blue}{Note: This paper represent a part of early result of our official world-action model zero-shot sim-to-real transfer work, which will be released soon.}
\end{abstract}
\vspace{-4pt}
 
\section{Motivation}
\vspace{-2pt}
 
Training manipulation policies in the real world is prohibitively expensive. While simulation offers a scalable alternative through parallelized data collection~\cite{isaac_gym, zhao2020sim2real}, inherent visual and physical discrepancies, known as the sim-to-real gap, often lead to transfer failures~\cite{tobin2017domain, openai2019rubiks}. A successful sim-to-real system can therefore reduce or eliminate the need for costly real-world teleoperation data.
 
We focus on \emph{world-action models}: unified generative models that predict future visual observations together with robot actions~\cite{ha2018world, du2024video, cosmos_policy}. Prior work has demonstrated the promise of such models in sim-to-sim or real-to-real settings, but it remains unclear whether they can cross the simulation-to-real gap for robotic manipulation. This gap is important because a positive result would extend the benefits of sim-to-real learning---cheap, scalable, automatically generated data---to a new class of policy models.
 
Our contribution is therefore not an ablation proving that joint video-action prediction is necessary for sim-to-real transfer. Instead, we ask a simpler empirical question: \emph{can a world-action model trained entirely from synthetic demonstrations perform zero-shot real-world manipulation?} To answer this question, we combine Cosmos Policy with extensive domain randomization and AnyTask's~\cite{anytask} automated demonstration generation, and present, to our knowledge, the \emph{first successful sim-to-real transfer of a world-action model}.

\vspace{-2pt}
\section{Methodologies \& Experiments}
\vspace{-2pt}

\subsection{Cosmos Policy}
Cosmos Policy~\cite{cosmos_policy} adapts a pretrained video diffusion model (Cosmos-Predict2) into a robot policy through single-stage post-training. In this framework, actions, future observations, and value estimates are encoded as latent frames within a unified diffusion process.
 
\vspace{-2pt}
\subsection{Simulation and Demonstration Generation}
\vspace{-2pt}
 
We construct training environments using GPU-accelerated simulation~\cite{isaac_gym} with comprehensive domain randomization:
(1)~\emph{textures}: object surfaces, table materials, and backgrounds drawn from procedural and photographic libraries;
(2)~\emph{camera poses}: perturbed translations and rotations of wrist and third-person cameras;
(3)~\emph{lighting}: varied number, position, intensity, and color temperature of scene lights;
(4)~\emph{object placements}: uniformly sampled within the reachable workspace.
These variations are designed to cover a broad range of visual appearances so that the real workspace is more likely to fall within the randomized training distribution, without requiring us to manually match the real test scene~\cite{tobin2017domain}.
 
We leverage AnyTask~\cite{anytask} to generate synthetic priors, i.e., expert demonstrations, via foundation-model-guided motion planning (ViPR) without human teleoperation across four tasks (\textit{lift banana}, \textit{lift brick}, \textit{open drawer}, \textit{put strawberry into bowl}), as shown in Figs.~\ref{fig:title} and~\ref{fig:pred_effect}. This process produces ${\sim}$\textbf{800 demonstrations} per task (${\sim}3{,}200$ total) of RGB observations paired with end-effector action trajectories.
 
\vspace{-2pt}
\subsection{World-Action Model Training}
\vspace{-2pt}
 
The training of Cosmos Policy runs for \textbf{32} epochs on \textbf{40 H100} GPUs (${\sim}$\textbf{72} hours).
Figure~\ref{fig:pred_effect} provides a qualitative forward-prediction check on real executions: the model-predicted frames are visually consistent with the synchronized real camera observations at the approach, grasp, and lift stages. This visualization suggests that the model has learned real-scene dynamics useful for execution, even though all policy training data come from simulation.
 
\vspace{-2pt}
\subsection{Zero-Shot Sim-to-Real Transfer Results}
\vspace{-2pt}

We deploy the trained policy directly on a real Franka Research 3 without collecting real demonstrations or fine-tuning the policy on real data. The only real-world setup is the standard camera configuration used at inference time: fixed RGB views from the wrist and third-person cameras are mounted and streamed to the policy. Table~\ref{tab:results} summarizes the success rates across the four tasks.

\begin{table}[h]
\vspace{-6pt}
\centering
\caption{Real-robot success rate over 10 trials per task. Our world-action model is trained with ${\sim}800$ synthetic demonstrations per task and \emph{zero} real demonstrations. Diffusion Policy (DP) baselines are trained with 10 or 50 real demonstrations per task, providing a real-data reference rather than an architecture-matched ablation.}
\label{tab:results}
\vspace{2pt}
\small
\setlength{\tabcolsep}{2.5pt}
\begin{tabular}{lccccc}
\toprule
\textbf{Method} & \textbf{Banana} & \textbf{Brick} & \textbf{Drawer} & \textbf{Strawb.} & \textbf{Avg.} \\
\midrule
DP w/ 10 real demos  & 0/10 & 0/10 & 2/10 & 0/10 & \textbf{5\%} \\
DP w/ 50 real demos  & 4/10 & 3/10 & \textbf{3/10} & 0/10 & \textbf{25\%} \\
\midrule
\textbf{Ours} (800 sim) & \textbf{5/10} & \textbf{5/10} &  2/10 & \textbf{2/10} & \textbf{35\%} \\
\bottomrule
\end{tabular}
\vspace{8pt}
\end{table}

\begin{figure}[h]
\vspace{-10pt}
\centering
\includegraphics[width=1.0\linewidth]{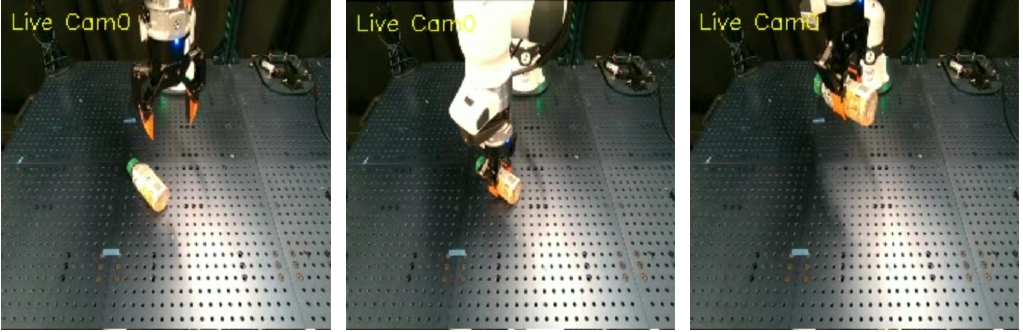}
\vspace{-8pt}
\caption{Out-of-distribution generalization on a real robot: although the bottle object is not part of the training task set, the policy approaches, grasps, and lifts it using the same zero-shot sim-to-real policy.}
\label{fig:novel_obj}
\vspace{-8pt}
\end{figure}

Our trained model achieves a~\textbf{35\%} average success rate using ${\sim}800$ fully synthesized simulation demonstrations per task and~\emph{zero} real-world demonstrations. The DP baselines in Table~\ref{tab:results} should not be interpreted as a controlled proof that world-action models are categorically better than Diffusion Policy; they use real demonstrations and therefore answer a different question. Instead, they provide a practical reference for the cost that sim-to-real methods aim to avoid. The key result is that a world-action model trained entirely from automatically generated synthetic data can achieve nonzero zero-shot real-world manipulation performance, reaching a higher average success rate than these real-demo DP references in our setup.

We attribute this transfer ability to the combination of pretrained video priors, broad domain randomization, and the joint action-video prediction objective, while leaving a controlled ablation of each factor to future work. Qualitatively, Fig.~\ref{fig:pred_effect} further shows that, even without manually reproducing the real test scene in simulation, the model's predicted future frames remain aligned with the live robot rollout.
 
\noindent\textbf{Out-of-distribution generalization.}
We further evaluate lifting a novel bottle unseen during training (Fig.~\ref{fig:novel_obj}).
The policy successfully approaches, grasps, and lifts it, suggesting that the model acquires \emph{generalizable visuomotor primitives} rather than only memorizing the original task objects.
 
\vspace{-2pt}
\section{Conclusion}
\vspace{-2pt}
 
We presented, to our knowledge, the first zero-shot sim-to-real transfer of a world-action model for robotic manipulation. Across four real-world tasks, the policy achieves 35\% average success using ${\sim}800$ fully automated simulation demonstrations per task and zero real robot demonstrations. This result shows that world-action models can inherit the central benefit of sim-to-real learning---reducing dependence on expensive real data---and opens a data-efficient direction for transferring generative robot policies from synthetic priors to the real world.
 
{
    \small
    \bibliographystyle{ieeenat_fullname}
    \bibliography{ref}
}
 
\end{document}